\documentclass{article}

\usepackage[nonatbib,final]{nips_2017}

\usepackage[utf8]{inputenc} % allow utf-8 input
\usepackage[T1]{fontenc}    % use 8-bit T1 fonts
\usepackage{hyperref}       % hyperlinks
\usepackage{url}            % simple URL typesetting
\usepackage{booktabs}       % professional-quality tables
\usepackage{amsfonts}       % blackboard math symbols
\usepackage{nicefrac}       % compact symbols for 1/2, etc.
\usepackage{microtype}      % microtypography
\usepackage{color}
\usepackage{graphicx}
\usepackage{amsthm}
\usepackage{cite}
\usepackage{titlesec}
\usepackage{tikzpagenodes}
\usepackage{wrapfig}
\usepackage{afterpage}
\usepackage{caption}
\usepackage{mdframed}

\titlespacing*{\section}{0pt}{0.7ex plus .2ex}{0.7ex plus .2ex}
\title{Spatial Uncertainty Sampling for End to End Control}% Learning for Self-Driving Cars}

\author{
  Alexander Amini \\
  CSAIL, MIT \\
  Cambridge, MA 02139 \\
  \texttt{amini@mit.edu}\\
  \And
  Ava Soleimany \\
  Biophysics, Harvard\\
  Cambridge, MA 02139 \\
  \texttt{soleimany@g.harvard.edu}\\
  \And
  Sertac Karaman \\
  LIDS, MIT \\
  Cambridge, MA 02139 \\
  \texttt{sertac@mit.edu}\\
  \And
  Daniela Rus \\
  CSAIL, MIT \\
  Cambridge, MA 02139 \\
  \texttt{rus@csail.mit.edu}
}

\begin{document}

\begin{tikzpicture}[remember picture,overlay,shift={(current page.north east)}]
\node[anchor=center,xshift=-10.5cm,yshift=-1.5cm,text width=14cm]{
\begin{center}
{\fontfamily{qcr}\selectfont
Originally Published in Neural Information Processing Systems (NIPS), Workshop on Bayesian Deep Learning 2017.
}
\end{center}
};
\end{tikzpicture}

\maketitle

\begin{abstract}
  %just jotting down ideas here: ....  Draw distinction from semantic segmentation and end to end control.

  % I would start the abstract with the problem - End-to-end trained neural networks are a compelling approach to autonmous vehicles because of their ..., but the challenging road conditions and ambiguous navigation situations require reliable uncertainty estimation.  Bayesian deep learning approaches to uncertainty estimation can be achieved with IID Gaussian priors over weights adn performing Monte Carlo drop out sampling at test time. However, ... In this paper, we propose an end-to-end autonomous driving system capable of uncertainty estimation, while preserving feature map corrleation. This advantage of our approach is better model fits, as well as tighter uncertainty estimate than ... We also demonstrate how uncerrtainties can be used in conjunctino with a human controller to ... We evaluate our algorithms on a challineng dataset collected over many different road types, times of day, and weather conditions.

  End-to-end trained neural networks (NNs) are a compelling approach to autonomous vehicle control because of their ability to learn complex tasks without manual engineering of rule-based decisions. However, challenging road conditions, ambiguous navigation situations, and safety considerations require reliable uncertainty estimation for the eventual adoption of full-scale autonomous vehicles. Bayesian deep learning approaches provide a way to estimate uncertainty by approximating the posterior distribution of weights given a set of training data. Dropout training in deep NNs approximates Bayesian inference in a deep Gaussian process and can thus be used to estimate model uncertainty. In this paper, we propose a Bayesian NN for end-to-end control that estimates uncertainty by exploiting feature map correlation during training. This approach achieves improved model fits, as well as tighter uncertainty estimates, than traditional element-wise dropout. We evaluate our algorithms on a challenging dataset collected over many different road types, times of day, and weather conditions, and demonstrate how uncertainties can be used in conjunction with a human controller in a parallel autonomous setting.

\end{abstract}

\section{Introduction}

End-to-end deep learning methods have yielded promising results in autonomous lane following \cite{gurghian2016deeplanes} and full end-to-end driving \cite{bojarski2016end, xu2016end}. Steering control outputs ($\mathbf{Y}$) can be directly learned from raw sensor data ($\mathbf{X}$) by finding a mapping $\mathbf{f}$ parameterized by weights $\mathbf{W}$ which minimizes some objective error. While powerful, these deep NNs are trained end-to-end as black boxes, and as such, lack a definitive measure of associated confidence with the output. Integration of these models into deployable or shared control frameworks necessitates a method for determining predictive uncertainty.

%Deep learning methods have demonstrated the ability to directly learn control outputs ($\mathbf{Y}$) from raw sensor data ($\mathbf{X}$) by finding a mapping $\mathbf{f}$ parameterized by weights $\mathbf{W}$ which minimizes some objective error. When applied to the autonomous driving task, these models, referred to as ``end-to-end'' (sensors to actuation), have yielded promising results in autonomous lane following \cite{gurghian2016deeplanes} and full end-to-end driving \cite{bojarski2016end, xu2016end}. While powerful, these deep NNs are trained end-to-end as black boxes, and as such, lack a definitive measure of associated confidence with the output. Integration of these models into higher level or shared control frameworks necessitates a method for determining predictive uncertainty.

Bayesian neural networks aim to learn a posterior distribution over the weights, $P(\mathbf{W} \vert \mathbf{X}, \mathbf{Y})$, and thus provide a grounded mathematical framework for determining model uncertainty. In practice, it is computationally infeasible to directly learn this posterior, $P(\mathbf{W} \vert \mathbf{X})$ from vast, and often noisy, observational data. Thus, variational inference (VI) methods have been used to obtain an approximation for the posterior to estimate model certainty. However, VI methods can also suffer from prohibitive computational cost \cite{paisley2012variational,kingma2013auto,rezende2014stochastic,hoffman2013stochastic}. Dropout sampling has emerged as an accurate, computationally efficient way to estimate model uncertainty. Applying dropout before every weight layer in a NN approximates a deep Gaussian process, since a dropout NN model minimizes the Kullback-Liebler divergence between the approximated posterior and the true posterior, $P(\mathbf{W}\vert \mathbf{X}, \mathbf{Y})$ \cite{gal2016dropout,damianou2013deep}. Thus, dropout sampling provides an estimation of the predictive variance.  Dropout, initially developed to avoid overfitting during training, places independent identically distributed Bernoulli random variables over every neuron to either ``drop'' or ``keep'' with some probability \cite{srivastava2014dropout}. This approach is problematic when individual learned feature maps exhibit strong correlation, such as in the preliminary network layers for computer vision applications. Spatial dropout overcomes this limitation and achieves uncertainty estimates while also promoting independence between convolutional feature maps \cite{gal2015bayesian,tompson2015efficient}.

%In recent years, deep learning has demonstrated the ability to directly learn control outputs from raw sensor data. These models, referred to as ``end-to-end'' (sensors to actuation), have yeilded promissing results in autonomous lane following \cite{}, and full end-to-end driving \cite{}. While powerful, these deep NNs are trained end-to-end as black-boxes, and as such, lack a definitive measure of associated confidence with the output and consequently cannot be integrated into a shared control (parallel autonomy) framework. While Bayesian NNs have been applied to certain aspects of the autonomous driving task, such as semantic segmentation \cite{} and depth estimation \cite{} this paper tackles the more complex task of learning an end to end model to control a full-scale autonomous vehicle. We formulate how our uncertainty estimations can be fused with human input thus enabling a novel parallel autonomy platform.

Here, we address the need for reliable uncertainty estimation in end to end learning systems for autonomous vehicles. The contributions of this paper can be summarized as follows: (1) Demonstration of accurate uncertainty estimation utilizing 2D spatial dropout; (2) A deep Bayesian network for end-to-end autonomous vehicle control; and (3) A shared control framework which incorporates uncertainty estimates, thus enabling shared human-machine interaction.

\section{Related Work}

The idea of using Bayesian integration over neural network parameters \cite{denker1987large,tishby1989consistent} and of averaging over the prior Gaussian weight distributions to estimate uncertainty of the output \cite{denker1991transforming, mackay1992practical, neal2012bayesian} have been widely explored. Recently, due to increase in training data sizes, more computationally tractable approaches have emerged. Specifically, it was shown that dropout \cite{srivastava2014dropout} can be used to estimate uncertainty by sampling the approximate underlying Gaussian process \cite{gal2016dropout, gal2016theoretically}. The effect of different dropout distributions (Bernoulli, Gaussian, and Spike-Slab) \cite{mcclure2016representing}, convolution specific dropout techniques \cite{gal2015bayesian}, as well as the impact of model versus data uncertainty \cite{kendall2017uncertainties} has also been explored.

This notion of stochastically sampling from Bayesian NNs to achieve uncertainty estimates has been applied to disease detection \cite{leibig2017leveraging}, reinforcement policy learning \cite{gal2016dropout}, as well as scene segmentation \cite{kendall2015bayesian} and depth estimation \cite{kendall2017uncertainties} for autonomous vehicles. Bayesian uncertainty has not previously been applied to the more challenging autonomous actuation aspect of end-to-end control \cite{bojarski2016end,xu2016end}, and it has not yet been shown how uncertainty can be incorporated into a parallel autonomy system \cite{naser2017parallel} where human and robot (i.e., vehicle steering system) share control. Our approach addresses these limitations by using spatial dropout \cite{gal2015bayesian} to estimate uncertainty while also promoting independence between feature maps during training of our end-to-end driving network.

\section{Methodology}
%\todo{ava, read this article if helpful: https://wiki.math.uwaterloo.ca/statwiki/index.php?title=dropout}

%
%
\begin{figure}[t!]
\includegraphics[width=\linewidth]{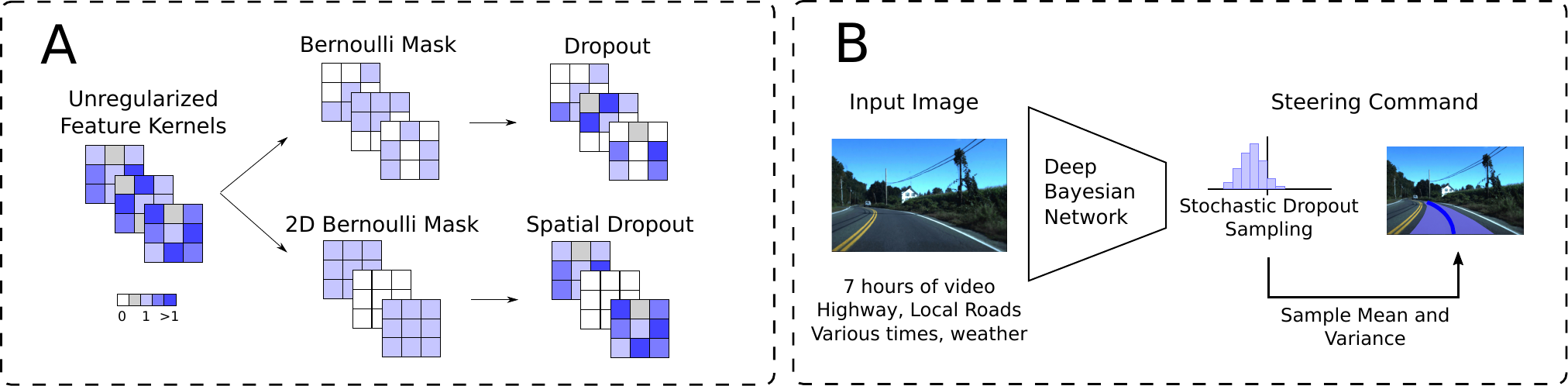}
\centering
\caption{\small {  \textbf{A. Dropout Comparison.} Element wise dropout (top) compared to 2D spatial dropout with Bernoulli random variables. \textbf{B. End-to-end uncertainty estimation for autonomous vehicle control.} Pipeline for training a deep Bayesian network to output stochastic samples of control, which in turn can be used to predict a single steering command and associated uncertainty. }}
\label{fig:network_and_spatial_dropout}
\end{figure}

We implement two dropout based approaches for estimating model uncertainty in a Bayesian NN for end-to-end autonomous vehicle control. It can be shown that applying dropout before every weight layer is equivalent to approximating a probabilistic deep Gaussian process \cite{gal2016dropout}. From the Dropout-based approximation of the posterior $q(\mathbf{W})$, we obtain a predictive distribution $q(\mathbf{Y}\vert\mathbf{X}) = \int P(\mathbf{Y}\vert \mathbf{X}, \mathbf{W})q(\mathbf{W})d\mathbf{W}$. Given $T$ stochastic forward passes through the network using dropout, $\{ \mathbf{W_t} \}_{t=1}^T$, we define the predictive mean as: $\mathbb{E}[\mathbf{Y} \vert \mathbf{X}] = \frac{1}{T}\sum_{t=0}^T \mathbf{f}(\mathbf{X},\mathbf{W_t})$ and define the model uncertainty (e.g., predictive variance) as $\textrm{Var}[\mathbf{Y} \vert \mathbf{X}] = \frac{1}{T}\sum_{t=0}^T \mathbf{f}(\mathbf{X},\mathbf{W_t})^2 - \mathbb{E}[\mathbf{Y} \vert \mathbf{X}]^2$.

%The predictive mean and variance of our model, taken over this predictive distribution, are given by $\mathbb{E}_{q(y|x)}[\mathbf{Y}]$ and $\textrm{Var}_{q(y|x)}[\mathbf{Y}]$.

In this paper, we use two methods for the aforementioned stochastic sampling: (1) element-wise Bernoulli dropout \cite{srivastava2014dropout} and (2) spatial Bernoulli dropout \cite{tompson2015efficient}, illustrated in Figure~\ref{fig:network_and_spatial_dropout}A. 
\begin{wrapfigure}{r}{0.4\textwidth}
  \centering
  \includegraphics[width=0.4\textwidth]{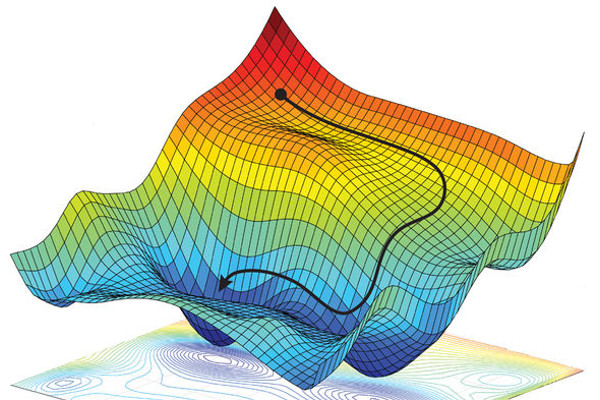}
  \caption{\small \textbf{Non-convex optimization}. We utilize stochastic gradient descent to find a local optimum in our loss landscape.}
  \label{fig:sgd}
  \vspace{-10pt}
\end{wrapfigure}
In the second approach, uncertainty estimates result from sampling a Bernoulli random variable $\mathbf{Z}$ to drop entire feature maps: $\mathbf{z}^{(k,l)} \sim Bernoulli(p)$ corresponds to $k$-th feature map in the $l$-th layer, where $p$ is the probability that all units in the feature map remain active \cite{gal2015bayesian}. Equivalently, this can be thought of as a special case of element-wise dropout, further supporting the notion that the sample variance from $T$ stochastic passes of spatial dropout provides a valid way to estimate the uncertainty. Specifically, we consider each neuron in a feature map to be drawn from a Bernoulli distribution $\mathbf{\widetilde z}^{(i,k,l)} \sim Bernoulli(p)$, corresponding to the $i$-th neuron in the $k$-th feature map from the $l$-th layer. Then, spatial dropout precisely corresponds to the instance where $\mathbf{\widetilde z}^{(i,k,l)}=\mathbf{\widetilde z}^{(j,k,l)} \, \forall (i,j)$ pairs across feature maps in convolutional layers. Because we can think of spatial dropout as a special instance of element wise dropout with an additional constraint, it follows that spatial dropout provides a valid way to estimate the model uncertainty. We implemented both approaches and compared the resulting loss optimization and uncertainty estimates.

We tested the efficacy of spatial dropout sampling of deep Bayesian networks applied to the complex task of end-to-end driving (Figure~\ref{fig:network_and_spatial_dropout}B). The vehicle base platform used for this study is a Toyota Prius 2015 V, which we retrofitted with sensors, power, and computing systems for parallel and autonomous driving. We collected an extensive dataset with over 7 hours and 500GB of driving data across a  range of road and weather conditions to train our models. All models were trained on the NVIDIA DGX-1 supercomputer and were able to run in real-time on the car's NVIDIA Drive PX2.

Our network takes as input a single image frame from a front facing video camera and performs feature extraction with 5 convolution layers with $5\times 5$ kernels and alternating ($2\times 2$, $1\times 1$) strides. The resulting features are then fed into 4 fully connected layers which perform dimensionality reduction until we get a single output unit at the last layer. Dropout is performed after every layer with $p=\{0.9,\, 0.5\}$, where $p$ corresponds to the probability of being kept, for convolutional and fully connected layers, respectively. The network was trained end-to-end using stochastic gradient descent (Fig.~\ref{fig:sgd}) to optimize the mean squared error of the inverse-turning radius of the vehicle (monotonic in steering wheel command).

\section{Experiments and Results}

\begin{figure}[b!]
\includegraphics[width=\linewidth]{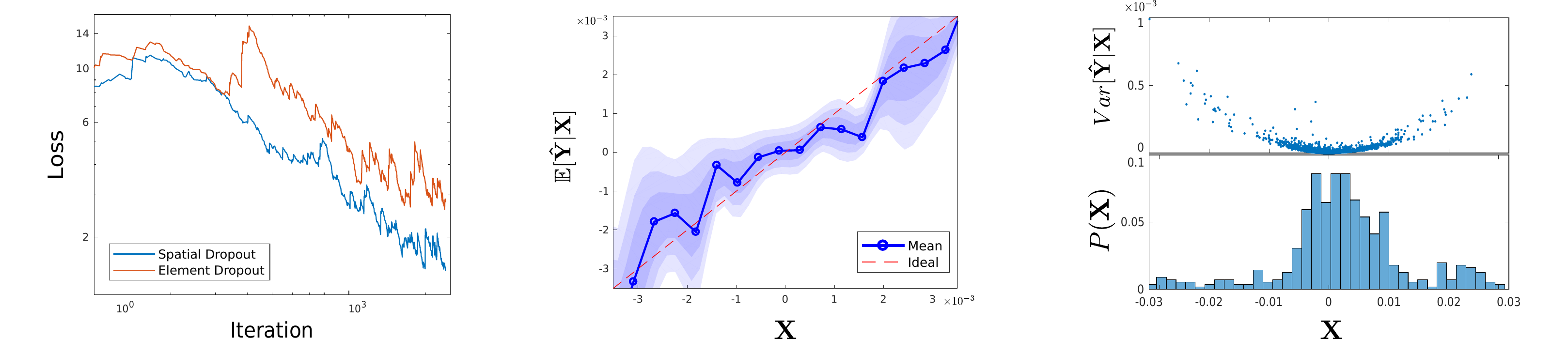}
\centering
\caption{\small{ \textbf{Results}. Comparison of the loss optimization using spatial and element wise dropout (left). Evaluation of the predictive mean (center) and predictive variance (right top) against the ground truth control, $\mathbf X$. The underlying data distribution $P(\mathbf X)$ is also illustrated on the bottom right. Note that, as expected, uncertainty (variance) increases as the amount of data decreases.
}}
\label{fig:results}
\end{figure}

In this section, we outline the uncertainty experiments performed during this study and introduce a novel control strategy to incorporate uncertainty estimates in a shared autonomous controller. We observed that utilizing spatial dropout during training  lead to faster convergence compared to element wise dropout (Figure~\ref{fig:results} left). After training the model, we were evaluated the predictive mean of new data as $\mathbb{E}(\mathbf{Y}\vert \mathbf{X})$ (center). The shades of blue represent half standard deviations from the mean and we observe a greater variance as we are given data farther from the origin. Specifically, this corresponds to having more variance in the predictive mean when given roads with tighter turns, as opposed to on relatively straight roads. Finally, we evaluate the predictive uncertainty, $\textrm{Var}(\mathbf{Y}\vert \mathbf{X})$ (right top), again as a function of the data. Note, that we observe a similar trend where uncertainty rises as the data moves farther from the origin (increased turns). As expected, we also observe that less data is available from these situations as well (right bottom) supporting a general trend where model uncertainty increases as the amount of data used during training decreases.

We also found that the mean uncertainty error, $\left( \frac{y-\mathbb{E}(\mathbf{Y}\vert \mathbf{X})}{\textrm{Var}(\mathbf{Y}\vert \mathbf{X})} \right)$, which penalizes estimators with high error and low uncertainty, for spatial dropout (1.74) was less than that obtained from element-wise dropout (2.67). Furthermore, we use these uncertainty estimates to develop a shared controller which fuses the human's actions with the network's output according to the uncertainty. In other words, we formulate a parallel autonomy control strategy $u_{PA} = (1-\sigma)\cdot u_N + \sigma \cdot u_H$ where $u_{PA}$, $u_{H}$, $u_{N}$ correspond to parallel autonomy, human, and network control commands for $\sigma = \kappa \textrm{Var}(\mathbf{Y}\vert \mathbf{X})$, with $\kappa$ being a tunable threshold parameter that depends on the drop probability. %and $\textrm{Var}(\mathbf{Y}\vert \mathbf{X})$ being the model uncertainty estimated from dropout sampling.

\section{Conclusion}
In this paper, we present a novel way to utilize spatial dropout to stochastically sample and compute uncertainty estimates of an end-to-end autonomous vehicle control system. We demonstrate how our estimates allow us to understand the limitations of our model when performing inference and thus provide a tunable framework which fuses model and human control. We are currently working to extend this work to not only estimate the posterior over a single control output but instead over a full parametric Gaussian Mixture Model (GMM) across all possible control outputs. In doing so, we will estimate the uncertainty associated to each of the individual GMM parameters and gain an even richer understanding of our model's confidence.

\section*{Acknowledgments}

Support for this work was given by the Toyota Research Institute (TRI).  However, note that this article solely reflects the opinions and conclusions of its authors and not TRI or any other Toyota entity. We gratefully acknowledge the support of NVIDIA Corporation with the donation of the DGX-1 used for this research.

\footnotesize
\bibliography{ref}

%\section*{Appendix}
%\todo{include model description and hyperparameters? results on devens with overlaid vis mask...}

\end{document}